\documentclass[letterpaper]{article} 
\usepackage[preprint]{aaai2027}  
\usepackage[hyphens]{url}  
\usepackage{graphicx} 
\usepackage{natbib}  
\usepackage{caption} 
\usepackage{algorithm}
\usepackage{algorithmic}

\usepackage{newfloat}
\usepackage{listings}
\DeclareCaptionStyle{ruled}{labelfont=normalfont,labelsep=colon,strut=off} 
\floatstyle{ruled}
\newfloat{listing}{tb}{lst}{}
\floatname{listing}{Listing}

\usepackage{booktabs}

\usepackage{arydshln}
\usepackage{placeins}

\title{Creative Product Design under Brand Image Constraints}
\author {
    Takeshi Onishi,
    Hiroya Makino,
    Takahiro Yamaguchi,
    Hiroyuki Sakai
}
\affiliations {
    TOYOTA CENTRAL R\&D LABS., INC.\\
    \{Takeshi.Onishi, hirom, t-yamaguchi, sakai\}@mosk.tytlabs.co.jp
}

\begin{document}

\maketitle

\begin{abstract}
Industrial design requires creative visual exploration while satisfying constraints that are often product-specific and implicit. Existing text-to-image (T2I) approaches can generate creative designs through concept blending or multi-agent self-enhancement, but they provide limited mechanisms for ensuring that generated designs remain consistent with such constraints. We focus on brand image as a representative design constraint and propose a framework for brand-aware redesign generation. Given a product name, the framework constructs a brand image survey report that summarizes product and market information, derives a creative direction from the report, and integrates this direction into a self-enhancement loop in which generated images are critiqued and prompts are revised. We evaluate the framework on 50 redesign tasks across five product categories, comparing it with Naive T2I, IP-Adapter/IT-Blender, and unconstrained self-enhancement under the same T2I backbone. By a VLM-as-Judge protocol to measure creativity and brand identifiability, the proposed framework achieves a more favorable trade-off between the two objectives, with particularly clear gains in highly constrained categories such as automotive and motorcycle, where the plausible design space is relatively limited compared with that of other product categories. These results suggest that dynamically eliciting and textualizing brand image can guide creative product design generation under brand image constraints.
\end{abstract}

\section{Introduction} \label{sec:intro}
In industrial design, a product's distinctive appearance plays a key role in differentiating it in the market and communicating its value to customers; hence, designing such an appearance requires not only visual originality but also compliance with various industrial constraints. This requirement distinguishes industrial design from artistic creation. For example, a sculpture may take any shape as long as it remains physically stable, whereas a vehicle design must include essential components such as doors and tires and must avoid sharp exterior features to ensure pedestrian safety. Beyond such functional and safety requirements, a product is also often expected to convey the visual identity of its brand. As a result, industrial designers must explore a large design space under strict constraints. This process depends heavily on expert intuition and experience, imposing a substantial burden on designers.

To reduce this burden and improve the efficiency of the design process, text-to-image (T2I) models have recently been actively studied~\cite{liu2023application,barros2024designing}. Prior work has mainly introduced creativity by optimizing concept embeddings, blending internal representations, or exploring prompts with multi-agent systems. ConceptLab~\cite{richardson2023conceptlabcreativeconceptgeneration}, for instance, aims to generate creative images by learning vector representations that intentionally depart from known concepts. IP-Adapter~\cite{ye2023ipadaptertextcompatibleimage} and IT-Blender~\cite{cho2025imagine} synthesize new concepts by blending the vector representations of multiple concepts within a neural network. CREA~\cite{venkatesh2025crea} introduces creativity by iteratively generating and evaluating images through a multi-agent self-enhancement process without task-specific training data.

\begin{figure}[t]
    \centering
    \includegraphics[width=0.40\textwidth]{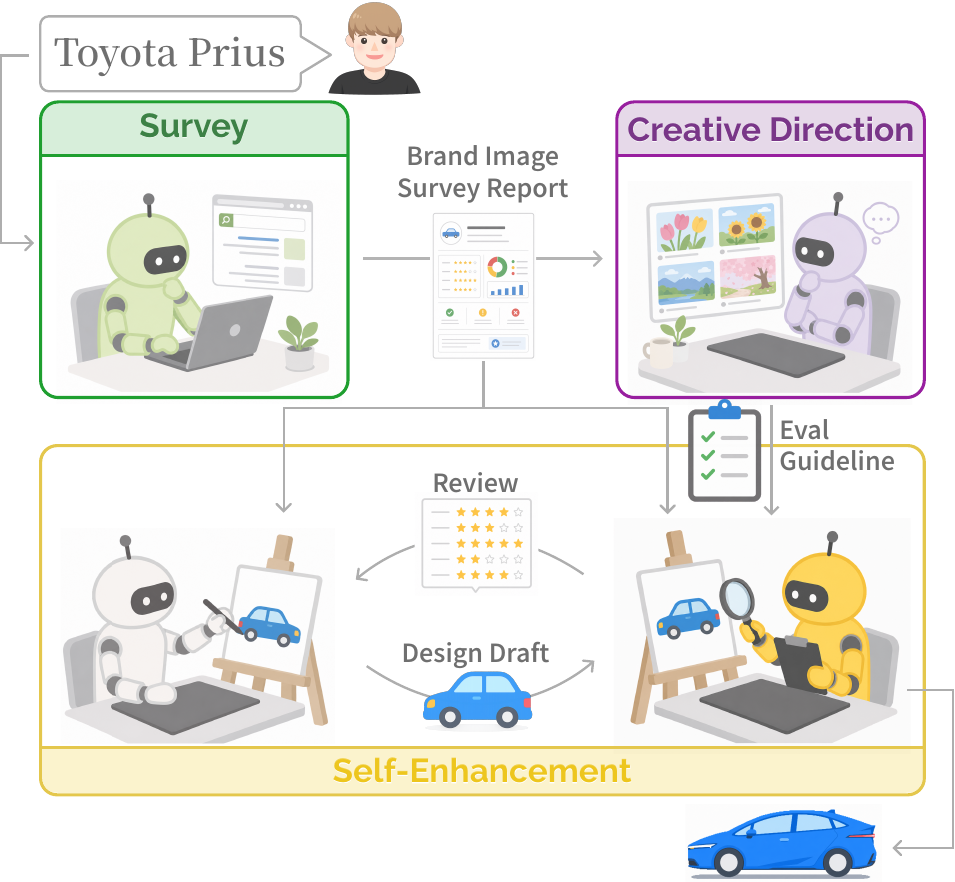}
    \caption{Overview of the proposed framework.}
    \label{fig:framework_overview}
\end{figure}


However, these studies primarily focus on injecting creativity and rarely address whether the resulting images remain acceptable under industrial design constraints. As a result, generated images may appear visually creative while violating requirements that real products must satisfy. This mismatch limits the practical applicability of existing generative approaches to industrial design.

In this study, we focus on brand image---the recognizable impression associated with a brand---as a representative industrial design constraint because it directly affects the appearance of products across a wide range of categories. Indeed, an industry survey reports that 86\% of companies consider brand consistency important~\cite{DemandMetric2016BrandConsistency}. Practical product design therefore requires creative visual exploration while preserving the product's brand image. Existing T2I systems struggle to satisfy this requirement because brand image is often implicit and product-specific, rather than a fixed constraint that can be specified once in a generic prompt or encoded in model parameters.

To explore creative product designs under brand image constraints, we propose a multi-agent framework. Given a product name, the framework handles the implicit and product-specific nature of brand image by retrieving product information from the web and automatically converting it into a textual description called a brand image survey report. It then derives a creative direction by comparing the report with auxiliary inspiration images and performs self-enhancement guided by that direction under brand image constraints. By representing brand image as an explicit textual description rather than embedding it as a fixed latent representation or learning it from product-specific training data, the framework can flexibly update the constraint and adapt it to each product's brand image.

To evaluate the performance of the proposed framework, we formalize the task of brand-aware redesign generation, in which a system takes an existing product name and generates an image of a new design. Then, we score each generated image in terms of creativity and brand identifiability, as defined in the Evaluation Protocol section.
We evaluate the framework on redesign generation for ten products in each of five categories (50 products in total). To improve reproducibility, we adopt a VLM-as-Judge protocol, in which a vision-language model (VLM) performs the scoring. The results show that the framework generates highly creative designs more reliably than strong baselines while better preserving brand image.

\section{Related Work} \label{sec:related_work}

Our target task, brand-aware redesign generation, can be regarded as a special case of T2I generation in which the system takes a product name as input and produces a creative design for that product while preserving its brand image. Two major lines of T2I research are relevant to this task: instruction-following methods, which generate images from detailed user specifications, and creativity-oriented methods, which generate images from abstract design goals.

\subsection{Instruction-Following T2I Models}

Instruction-following T2I models constitute a major line of T2I research that treats generation as the problem of faithfully reproducing user-specified content~\cite{ramesh2022hierarchical,saharia2022photorealistic,rombach2022high}. Several extensions enable conditioning beyond plain text prompts, including instruction-based image editing~\cite{brooks2023instructpix2pix}, image-conditioned generation~\cite{ruiz2023dreambooth}, and structure-conditioned generation~\cite{zhang2023controlnet,mou2024t2iadapter}. Because these approaches expect the user to specify the visual details to be rendered, they are suboptimal for our target task, in which user instructions are abstract and under-specified (e.g., ``redesign the Toyota Prius''). In contrast, the proposed framework does not require the user to specify creative visual details: it retrieves product-specific information from the web, summarizes it into a brand image survey report, and uses this report as a guide to autonomously explore new creative designs.

\subsection{Creativity-Oriented Image Generation}

Our work is more closely related to creativity-oriented image generation, which aims to produce creative images without detailed user specifications. A variety of approaches have been proposed to inject creativity into T2I generation.

ConceptLab~\cite{richardson2023conceptlabcreativeconceptgeneration} searches for a new concept under a given abstract category, such as ``pet'', while steering away from familiar instances such as ``dog'' or ``cat''. Because it locates a point in the latent space that is close to the abstract category but far from its familiar concrete instances, it requires a predefined set of such instances and is therefore designed for relatively well-defined abstract categories such as ``pet'' or ``vehicle''. This formulation is difficult to apply to a concrete target such as ``Toyota Prius'', which is the setting we consider, because neither a clear abstract category nor a set of guiding concrete instances is readily available.

Attention-based blending methods and multi-agent frameworks are more flexible with respect to the target concept. IP-Adapter and IT-Blender~\cite{ye2023ipadaptertextcompatibleimage,cho2025imagine} are attention-based blending approaches that introduce creativity by combining a textual concept and a reference image through learned attention. Another approach to combining a pair of images was proposed by \citet{makino2025zeroshotvisualconceptblending}. CREA~\cite{venkatesh2025crea}, a multi-agent framework, is arguably the closest related work to ours: it introduces creativity by iteratively generating, critiquing, and revising images. While these systems can introduce visual details that the user does not specify, they lack a mechanism for constraining the creativity introduced, such as preserving brand image, which is crucial in practical industrial design tasks. We demonstrate this limitation in our experiments, where these approaches fail to preserve the brand image of the original product, whereas the proposed framework successfully maintains it.

\section{Proposed Multi-Agent Framework} \label{sec:methodology}

To generate creative designs while preserving brand image, the proposed multi-agent framework extends self-enhancement~\cite{venkatesh2025crea} with two additional components: (1) a survey module that constructs a product-specific brand image survey report, and (2) a creative direction module that converts that report and auxiliary inspiration images into explicit evaluation criteria. The survey module makes the brand image explicit in textual form, whereas the creative direction module identifies a direction consistent with that brand image. Together, these two components make design constraints actionable throughout the iterative search process in the self-enhancement module. Figure~\ref{fig:framework_overview} summarizes the overall procedure implemented in the proposed framework.

\subsection{Survey Module}
The survey module takes a product name as input and produces a brand image survey report, a natural language document that summarizes the product, the market environment, the target customer, and the expected customer value. Although a user could provide an existing product report, we generate a report for each product in four steps using a large language model (LLM) agent named the Marketer to ensure a fair evaluation. First, Deep Research\footnote{https://openai.com/index/introducing-deep-research} gathers product details and reviews. Second, the LLM analyzes the market environment and competing products. Third, the LLM identifies a target customer profile from the product and market information. Fourth, the LLM summarizes the core value expected by that target customer. These outputs are then compressed into a short summary used in downstream prompting.

\begin{enumerate}
    \item Product details: What kind of product is it?
    \item Market environment: What market environment surrounds this product, including competing or similar products?
    \item Target customer: What are the primary customer demographics?
    \item Product key value: What core value does this product offer to customers?
\end{enumerate}
The specific prompt templates are given in the appendix.

\subsection{Creative Direction Module}
The creative direction module receives the survey report from the survey module and proposes a creative direction: a textual design strategy that specifies how creativity should be introduced while remaining consistent with the brand image described in the report. Then, this direction guides the self-enhancement process by extending the evaluation criteria of the Reviewer.

Inspired by mood boards~\cite{https://doi.org/10.1111/1468-5949.00250}, collections of images that stimulate designers' creativity, this module consists of a VLM-based agent, called the Brand Manager, and a set of auxiliary inspiration images. Each inspiration image serves as a seed for a candidate creative direction. Specifically, for each inspiration image, the Brand Manager first extracts the concept or tone of the image, then describes how that concept can be integrated into the target product without losing the product's essential identity, and finally writes a direction-specific evaluation guideline. The Brand Manager then compares the candidate creative directions obtained from all inspiration images and selects the one that is expected to most effectively appeal to the target customer identified in the survey report. In our experiments, we use 20 images randomly sampled from the web\footnote{https://randompicturegenerator.com/picture.php}. The corresponding prompt templates are given in the appendix.

The selected creative direction is passed to the self-enhancement module in the form of an evaluation guideline and is added to the evaluation criteria of the Reviewer. Consequently, the Reviewer judges generated images not only by generic creativity but also by how well they realize this direction. When an image receives a low score on this criterion, the accompanying comments are transformed into prompt-revision requirements in the next iteration. This design creates a direct link between high-level creative intent and low-level prompt updates, enabling self-enhancement to optimize not for creativity alone, but for creativity that remains aligned with brand image.

\subsection{Self-Enhancement Module}
Following CREA~\cite{venkatesh2025crea}, we implement the self-enhancement module, which improves generated images through iterative interaction between a Designer agent and a Reviewer agent.
The Designer writes prompts for the T2I model and generates candidate images. This process is inherently noisy because both prompt generation and T2I rendering are stochastic. Therefore, the module does not rely on a single generation step. Instead, it repeatedly revises the prompt based on the Reviewer's feedback and regenerates the image.

The Reviewer evaluates each generated image using multiple criteria, including the creative direction produced by the Brand Manager. The criteria are as follows:
\begin{itemize}
    \item Format: Does the image meet the format of the target product (e.g., no background)?
    \item Originality: How novel or unique is the image compared with existing designs?
    \item Unexpectedness: How much does the image combine elements in unconventional ways?
    \item Customer-value alignment: How well does the image align with the expected customer value identified in the survey report?
    \item Creative direction: How well does the image realize the chosen creative direction from the Brand Manager?
\end{itemize}
The Reviewer outputs both scores and textual comments for these criteria. The Designer then rewrites the prompt to address the low-scoring points, and the process repeats for a fixed number of iterations while retaining the best-scoring image and prompt obtained so far. The corresponding prompt templates are given in the appendix.

\section{Experiments} \label{sec:experiment}
\begin{table*}[htbp]
\centering

\begin{tabular}{p{0.95\textwidth}}
\hline
Category: Product Names \\
\hline

\texttt{automotive :} 
\small{Toyota Prius, Volkswagen Golf, Ford F-150, Honda Civic, Nissan Skyline, BMW X5, Mercedes-Benz C-Class, Tesla Model 3, Porsche 911, Audi A4} \\ \hdashline

\texttt{motorcycle :} 
\small{Honda Super Cub, Hero MotoCorp Splendor, Yamaha YZF-R Series, TVS Motor Apache, Bajaj Auto Pulsar, Suzuki GSX-R Series, Royal Enfield Classic, BMW R nineT, Harley-Davidson Sportster, KTM 690 Duke} \\ \hdashline

\texttt{beverage :} 
\small{Coca-Cola, Pepsi, Sprite, Fanta, Red Bull, Monster Energy, Gatorade, Pocari Sweat, Budweiser, Heineken} \\ \hdashline

\texttt{character\_series :} 
\small{Star Wars, Pokémon, One Piece, Dragon Ball, Harry Potter, Transformers, The Simpsons, Toy Story, Teenage Mutant Ninja Turtles, Super Mario} \\ \hdashline

\texttt{automobile\_manufacturer :} 
\small{Toyota Motor Corporation, Volkswagen AG, Ford Motor Company, Honda Motor Co., Ltd., Nissan Motor Co., Ltd., Bayerische Motoren Werke AG (BMW), Mercedes-Benz Group AG, Tesla, Inc., Dr. Ing. h.c. F. Porsche AG, Audi AG} \\

\hline
\end{tabular}

\caption{List of product names employed for the experiment. (5 categories $\times$ 10 individual products)} \label{table:product_names}

\end{table*}

To evaluate the proposed framework in practical industrial design scenarios, we use a controlled redesign setting that simulates the redesign of real industrial products. The benchmark spans product categories with different types and strengths of design constraints. We compare the proposed framework with baselines under a shared T2I backbone and assess all outputs using a common VLM-based evaluation protocol (VLM-as-Judge). The details of each step are described below.

\subsection{Target Products}
We evaluate the framework on 50 redesign tasks across five product categories. Each task requires generating a redesigned image of an existing product from its name and a set of candidate inspiration images. The goal is to produce a design that is both creative and consistent with the target brand image. The categories are automobiles (\texttt{automotive}), motorcycles (\texttt{motorcycle}), beverage bottles (\texttt{beverage}), character series (\texttt{character\_series}), and automobile manufacturer logos (\texttt{automobile\_manufacturer}).

These five categories were chosen to cover a range of constraint types and strengths. \texttt{automotive} and \texttt{motorcycle} are the most strongly constrained categories in our benchmark. Their designs are shaped by practical, ergonomic, and regulatory requirements, so the admissible design space is narrow and major visual changes must still preserve product identity. We treat these categories as model-change scenarios in which the product name and brand image remain fixed while the exterior is redesigned. \texttt{beverage} is also constrained by manufacturability and use, but the visual constraint is weaker because much of the variation can be expressed through labels, graphics, and surface treatment. This category therefore allows more room for visual experimentation while still requiring a plausible product form. \texttt{character\_series} differs qualitatively from these product-centered cases. Here the main constraint is not physical feasibility but consistency with a shared fictional world, visual language, and audience expectation. The constraint is therefore broad but implicit, requiring contextual understanding rather than compliance with explicitly specified rules. \texttt{automobile\_manufacturer} is constrained in yet another way. Because logos rely on a small set of shapes, symbols, and typographic cues, the design space is extremely compact, while the requirement to preserve brand image remains strong. This makes logo redesign a high-constraint setting in which the constraints are primarily semantic rather than geometric. The full target list is shown in Table~\ref{table:product_names}.

\subsection{Baselines}
We use baselines that introduce creativity into generated images through different mechanisms to evaluate whether the proposed framework improves the practical balance between creativity and brand image consistency in industrial design. Specifically, we compare against direct prompt-based generation, multimodal concept blending, and iterative creativity-oriented refinement. For a fair comparison, all approaches use the same T2I backbone, Flux.1-dev~\cite{greenberg2025demystifyingfluxarchitecture}. This comparison isolates how different creativity-injection strategies affect the trade-off between creativity and brand identifiability.

\textbf{Naive T2I:} This baseline serves as the minimal reference point without any explicit creativity-injection mechanism. It uses only the product name and a brief redesign instruction as the input prompt to the T2I model. Specifically, each product is generated from a prompt such as ``Newly designed \texttt{\{product name\}}'' without any survey stage, external inspiration images, or iterative prompt refinement.

\textbf{IP-Adapter/IT-Blender:} These baselines represent creativity injection through visual concept blending. For each product, we use a minimal text prompt similar to the one above together with an inspiration image, allowing external inspiration images to influence the redesign. Because these methods do not include a mechanism for selecting the most appropriate inspiration image, we generate one output for each candidate inspiration image and report the average score across them.

\textbf{Self-Enhancement / CREA~\cite{venkatesh2025crea}:} This baseline represents creativity injection through self-enhancement. To avoid confounds from implementation differences, we construct it within the proposed framework by removing the survey module and creative direction module while keeping the self-enhancement loop. The baseline therefore receives only the product name as minimal task information and performs iterative prompt revision under fixed evaluation criteria.

\subsection{Implementation Details}
All LLM- and VLM-based agents in the framework are implemented with GPT-5-mini, except that the initial product research is delegated to Deep Research. The self-enhancement loop runs for 15 iterations per product. Note that the internal Reviewer (GPT-5-mini) and the evaluation judge (Gemini 2.5 Pro) belong to different model families, which mitigates the risk of the framework optimizing directly toward the judge's preferences.

\subsection{Evaluation Protocol} \label{sec:evaluation_protocol}

For reproducibility and scalability, we adopt a VLM-as-Judge protocol and evaluate all generated images with Gemini 2.5 Pro~\cite{comanici2025gemini25pushingfrontier}. 
Evaluating creativity and brand identifiability with human annotators is inherently challenging because such judgments are sensitive to individual background, regional context, and temporal trends. Although VLM-based evaluation is not perfectly aligned with human judgment~\cite{10.5555/3692070.3692324}, recent studies on multimodal automatic evaluation report rapid improvements in human alignment for T2I assessment and related multimodal judging tasks~\cite{TUNG2025105371, abe2024assessingaestheticevaluationcapabilities}. 

Following this line of work, we evaluate creativity and brand identifiability independently. The full prompts are provided in the appendix.

\subsubsection{Creativity Evaluation}
We assess creativity in terms of novelty and usefulness, following a standard view in creativity research~\cite{Mumford01072003}. For each generated image, the VLM assigns a score from 1 to 5 on both dimensions using a category-specific evaluation prompt:

\begin{itemize}
    \item \textbf{Novelty Score}: How visually different and unexpected the design is relative to the existing product.
    \item \textbf{Usefulness Score}: How plausible, coherent, and usable the design appears as a product.
\end{itemize}
The final Creativity Score is defined as the harmonic mean of the Novelty Score and Usefulness Score. We use the harmonic mean rather than the arithmetic mean because it penalizes one-sided designs that are highly novel but implausible, or plausible but insufficiently novel. This choice reflects the view that creativity is valuable in industrial design only when novelty remains compatible with practical design constraints.

\subsubsection{Brand Identifiability Evaluation}
We evaluate brand image consistency in terms of recognizability rather than rule-level compliance. Although brand image can sometimes be formalized as an internal design language or a set of design rules, the more practical question in product presentation and sales is whether users can recognize the intended brand from visual appearance alone. Concretely, we ask whether a viewer could plausibly infer the intended brand or product identity from the generated image alone. Accordingly, the VLM is asked to identify the manufacturer or product represented by each image.

We use answer accuracy as the Identity Score. An answer is counted as correct if it matches one of the predefined acceptable aliases for the target brand or product. For each category, we compute the mean accuracy across the 10 target products in that category.

\section{Results} \label{sec:result}

\begin{figure}[tbp]
    \centering
    \includegraphics[width=0.40\textwidth]{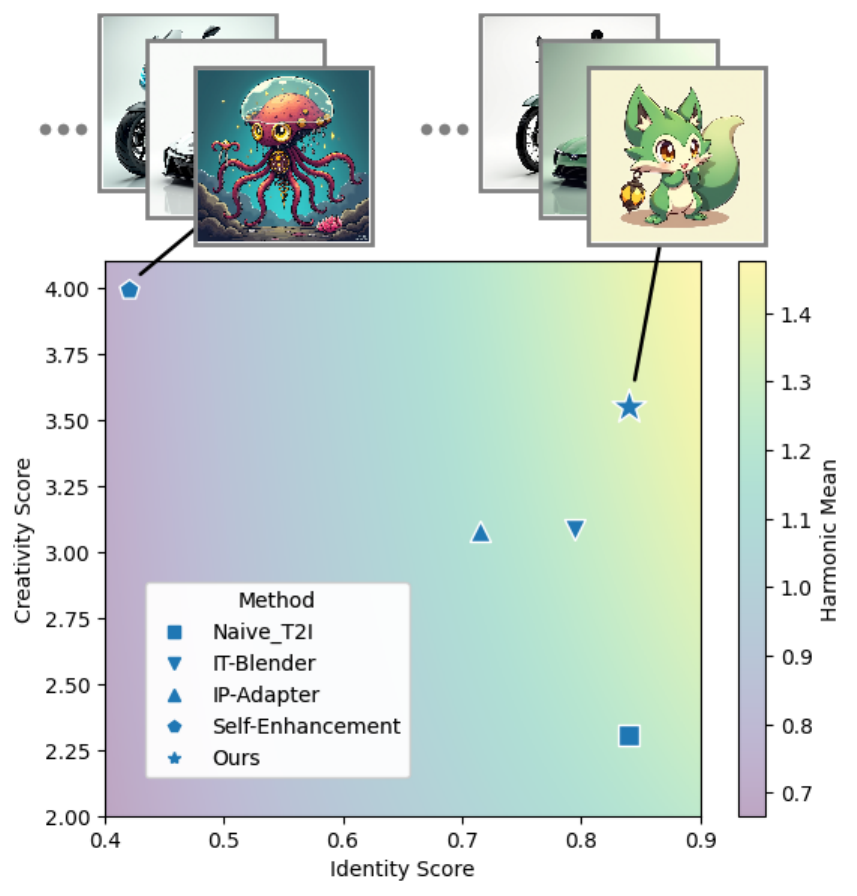}
    \caption{Creativity Score and Identity Score with example outputs from the proposed framework and Self-Enhancement.}
    \label{fig:main_score}
\end{figure}


Figure~\ref{fig:main_score} summarizes the Creativity and Identity Scores of the proposed framework and the baselines, along with example outputs from the proposed framework and Self-Enhancement.
The examples illustrate a qualitative difference between the methods: the proposed framework introduces creativity while preserving brand identifiability, whereas Self-Enhancement exhibits identity drift.
Overall, the quantitative results indicate that the proposed framework achieves a more favorable trade-off between these two objectives than the baselines.

\begin{figure*}[htbp]
    \centering
    \includegraphics[width=0.95\textwidth]{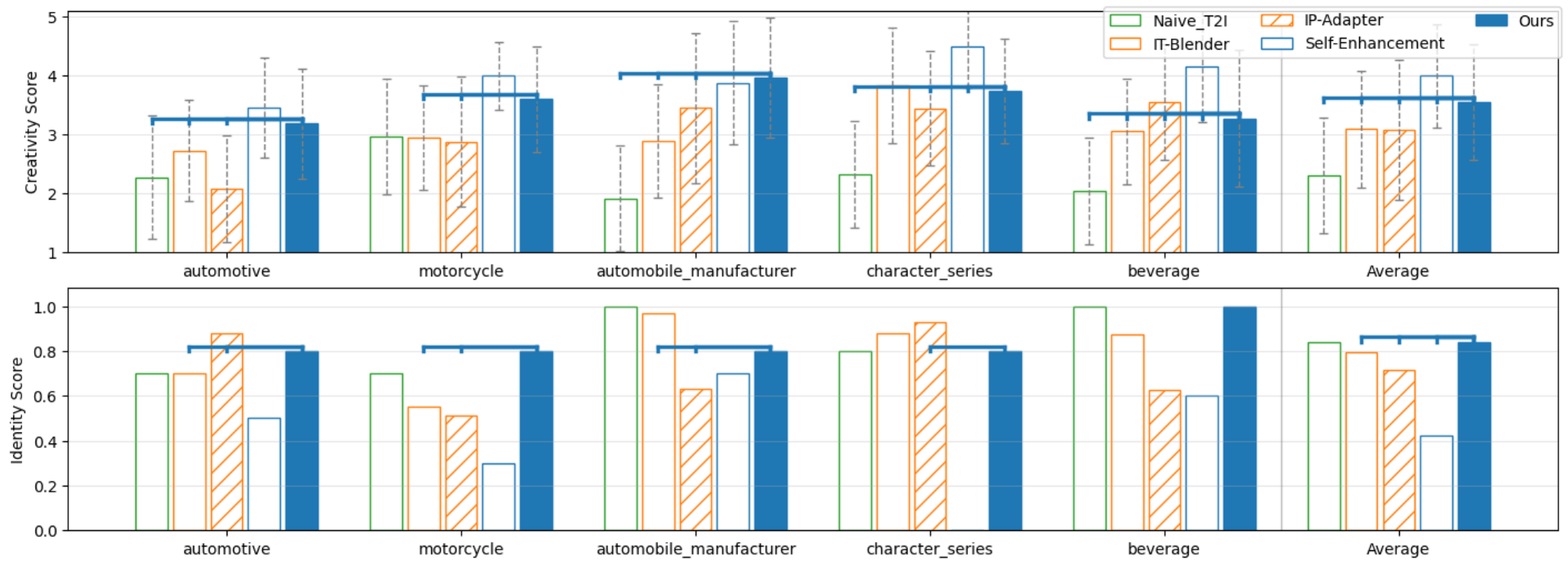}
    \caption{Creativity Score and Identity Score by category. Error bars on the Creativity Score indicate one standard deviation. Horizontal bars indicate statistically significant differences from the proposed framework, assessed by a t-test for the Creativity Score and McNemar's test for the Identity Score. Note that, as described in the Baselines subsection, IP-Adapter and IT-Blender generate one output for each candidate inspiration image (20 outputs per product), whereas Naive T2I and Self-Enhancement produce a single final image per product; consequently, the statistical power to detect significant differences is lower for these two methods.}
    \label{fig:main_score_by_cat}
\end{figure*}

Specifically, compared with IP-Adapter and IT-Blender, the proposed framework improves both the Creativity Score and the Identity Score.
Against Naive T2I, the proposed framework yields a higher Creativity Score while maintaining a comparable Identity Score.
In contrast to Self-Enhancement, which attains the highest Creativity Score, the proposed framework substantially improves the Identity Score while incurring only a moderate reduction in creativity.
These differences are statistically significant, as shown in Figure~\ref{fig:main_score_by_cat}.
Taken together, these results indicate that the proposed framework is more effective at balancing creativity and brand identifiability.

Figure~\ref{fig:main_score_by_cat} further breaks down the results by product category.
While the performance of the baselines varies across categories, the proposed framework performs robustly across all categories.

A closer inspection reveals that the advantage of the proposed framework is particularly pronounced in categories with strong structural or semantic constraints.
In \texttt{automotive} and \texttt{motorcycle}, all baselines struggle to simultaneously achieve high creativity and brand identifiability.
In particular, IP-Adapter, IT-Blender, and Naive T2I show degraded performance on both metrics relative to their performance in other categories, while Self-Enhancement improves creativity at the expense of substantial identifiability loss.
These results suggest that introducing creativity in these domains is inherently difficult because even small deviations can compromise critical brand-specific visual elements.
In contrast, the proposed framework maintains relatively high scores on both metrics, indicating that its creative direction module might help find a plausible creative direction that does not violate the target brand image.

\begin{figure}[h]
    \centering
    \includegraphics[width=0.95\columnwidth]{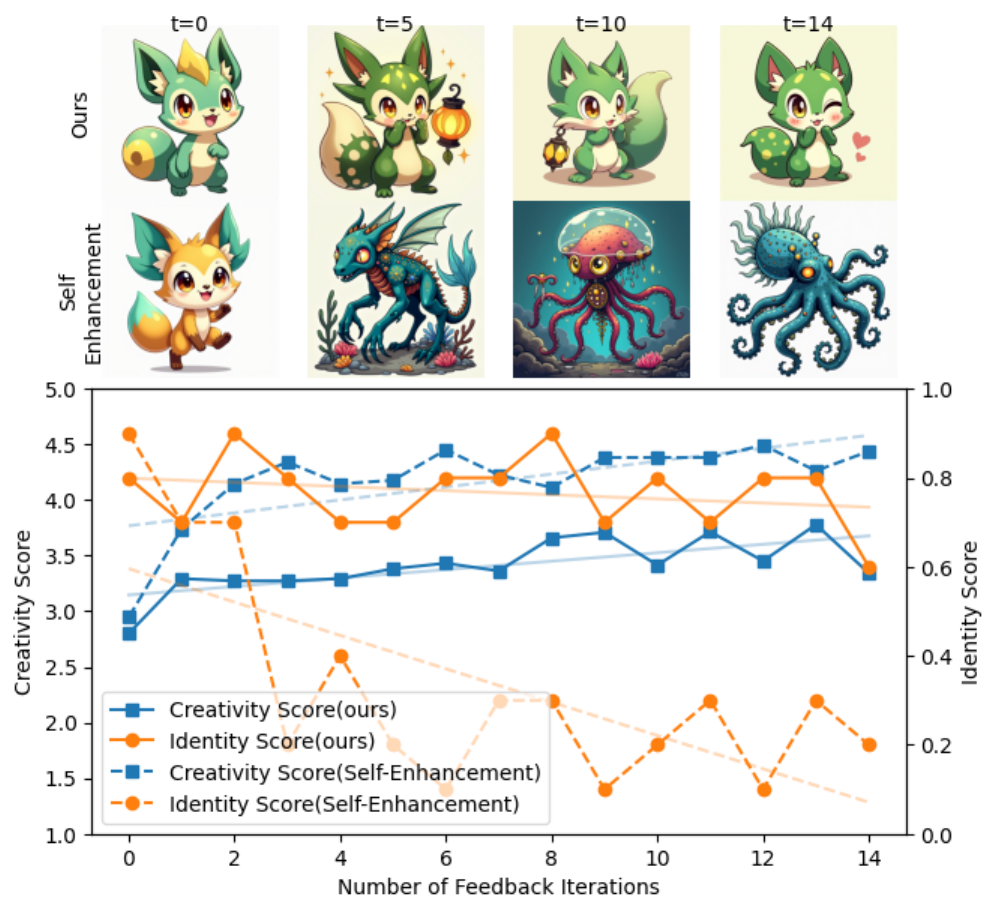}
    \caption{Creativity and Identity Scores during the self-enhancement loop in \texttt{character\_series}. Solid lines represent the proposed framework, and dashed lines represent Self-Enhancement. The images at the top show generated samples for Pok\'{e}mon at the corresponding iterations.}
    \label{fig:charactar_score}
\end{figure}

In \texttt{character\_series}, we observe a different failure mode.
Self-Enhancement achieves high creativity but substantially degrades brand identifiability compared with Naive T2I.
This suggests that the underlying T2I model already encodes implicit knowledge of character identity in its parameters, and that this knowledge is partially overwritten during the unconstrained iterative self-enhancement process.
Figure~\ref{fig:charactar_score} traces this process, showing that the Identity Score of Self-Enhancement declines over the iterative process with sample intermediate images of Pok\'{e}mon illustrating the identity drift.
In contrast, the proposed framework suppresses this identity drift: the Identity Score remains high throughout the loop while the Creativity Score gradually increases, and the framework ultimately attains an Identity Score of 0.8 and a Creativity Score of 3.7.
This result indicates that the brand image survey report, which explicitly encodes the visual and contextual characteristics of the series in textual form, plays a key role in stabilizing identity during the self-enhancement process.

In \texttt{automobile\_manufacturer} and \texttt{beverage}, all methods achieve relatively high Identity Scores.
This is likely because these categories contain strong and easily recognizable visual cues (e.g., typography or product names), which make brand identifiability less sensitive to design variation.
Even in these easier settings, however, the proposed framework remains competitive with or superior to the baselines overall, although its Creativity Score in \texttt{beverage} is slightly lower than that of some baselines.

Overall, the category-wise analysis highlights that the primary advantage of the proposed framework emerges in challenging settings where creativity and brand identifiability are in strong tension.
This supports our claim that explicitly modeling brand image as a dynamic textual representation is critical for practical design generation under industrial design constraints.


\subsection{Ablation Study} \label{sec:ablation}

\begin{figure}[t]
    \centering
    \includegraphics[width=0.40\textwidth]{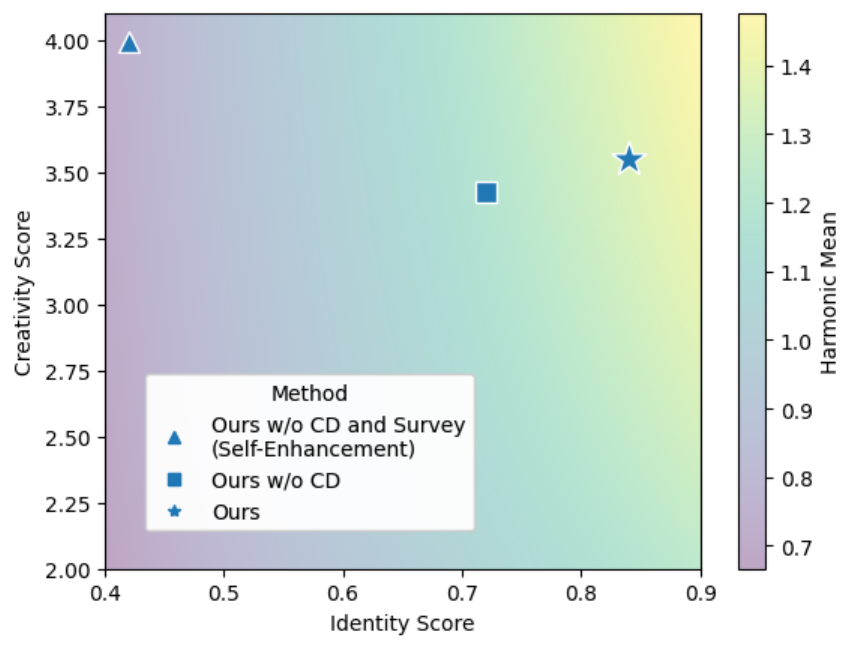}
    \caption{Ablation study of the proposed modules. \texttt{CD} denotes the creative direction module.}
    \label{fig:module_ablation}
\end{figure}


\begin{figure}[h]
    \centering
    \includegraphics[width=0.45\textwidth]{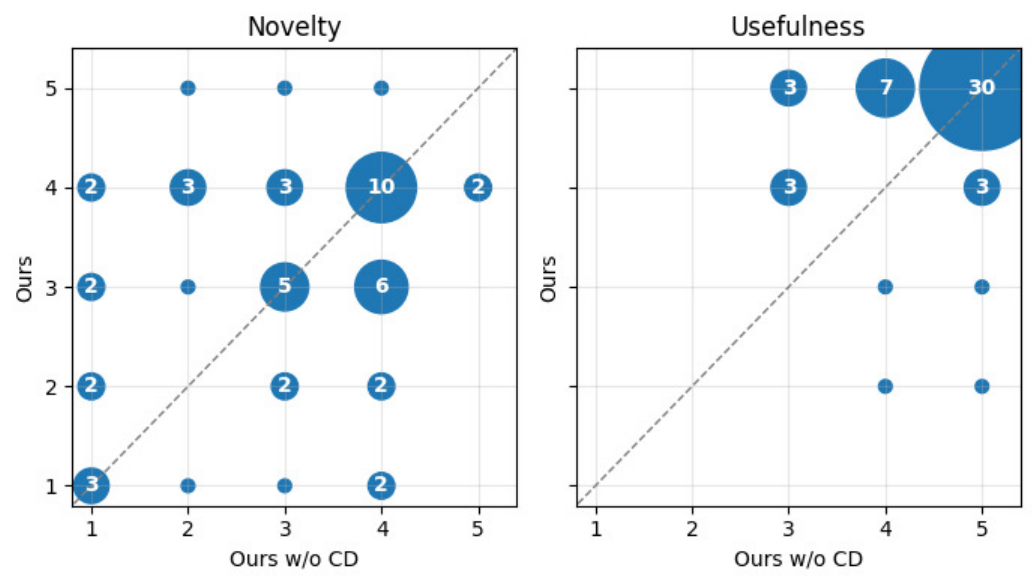}
    \caption{Novelty and Usefulness components of the Creativity Score for the proposed framework, \texttt{Ours}, and its ablated variant, \texttt{Ours w/o CD}. Each number in a circle indicates the number of products with identical score pairs.}
    \label{fig:creativity_ablation}
\end{figure}


In this section, we examine the contributions of the survey module and the creative direction module.
Figure~\ref{fig:module_ablation} compares the full framework, \texttt{Ours}, with two ablations: \texttt{Ours w/o CD}, which removes the creative direction module while retaining the survey report, and \texttt{Ours w/o CD and Survey}, which removes both proposed modules and is equivalent to Self-Enhancement.
Comparing \texttt{Ours w/o CD and Survey} with \texttt{Ours w/o CD} isolates the effect of the survey report.
The survey report substantially improves the Identity Score, but this improvement comes at the cost of a lower Creativity Score.
This result suggests that the survey report conditions the self-enhancement process to preserve brand identifiability.
Comparing \texttt{Ours w/o CD} with \texttt{Ours} further suggests that the creative direction module provides small, statistically non-significant gains in both scores.

To better characterize these gains, Figure~\ref{fig:creativity_ablation} decomposes the Creativity Score into Novelty and Usefulness Scores for \texttt{Ours} and \texttt{Ours w/o CD}.
Novelty does not show a consistent shift, whereas Usefulness improves for several lower-scoring cases and remains unchanged for many already high-scoring cases.
This pattern suggests that the creative direction module may primarily improve usefulness by helping the framework select more plausible creative directions from among multiple inspiration images.

\section{Limitations} \label{sec:limitation}
The first limitation concerns the evaluation protocol. Our evaluation relies on a VLM-as-Judge protocol rather than assessments by human designers, domain experts, or target customers. This protocol improves scalability and reproducibility and reduces reliance on human evaluation in settings where generated designs may raise copyright concerns, but cannot fully capture subjective creativity, consumer perception, or professional judgments of brand image preservation.

The second limitation is the scope of constraints considered. We consider only brand image, although industrial design also imposes safety, ergonomic, dimensional, mechanical, material, manufacturing, regulatory, and cost constraints. The proposed framework neither explicitly models these constraints nor verifies physical realizability. Supporting downstream engineering or production decisions requires handling multiple hard and soft constraints, potentially with priority-aware evaluation criteria.

\section{Conclusion} \label{sec:conclusion}

This study addresses a central challenge in applying generative models to industrial design: how to explore creative designs while preserving brand image as a normative constraint. Previous T2I-based approaches can generate creative appearances through concept blending or exploratory optimization, but they do not adequately account for the constraints required in real product design. As a result, their outputs often drift away from the intended product context. We therefore introduced a framework that treats brand image as an explicit, dynamically updated design constraint rather than as a fixed prompt or latent representation.

The proposed framework combines three components: (1) a survey module that uses an LLM and web search to summarize product-specific brand image and market context, (2) a creative direction module that uses the brand image survey report and inspiration images to define a creative direction and associated evaluation criteria, and (3) a self-enhancement module in which a Reviewer agent repeatedly critiques generated outputs and a Designer agent updates prompts accordingly. This design makes brand image actionable as an evaluable constraint throughout the search process and enables controlled trade-offs between creativity and brand identifiability.

Across 50 redesign tasks spanning five product categories, the proposed framework improved the balance between creativity and brand identifiability compared with Naive T2I, IP-Adapter/IT-Blender, and the Self-Enhancement baseline. The gains are particularly meaningful in the highly constrained automotive and motorcycle categories.

Overall, the results demonstrate that dynamically eliciting and textualizing brand image, then integrating it into iterative design exploration, is a practical way to reconcile creativity with constraint satisfaction. Although we focus on brand image, the framework can be extended to other normative constraints and may serve as a foundation for practical design-space exploration with T2I models.

\section{Ethical Statement} \label{sec:ethics}
All referenced trademarks, product designs, and characters belong to their respective owners; generated images are unofficial research artifacts used only for evaluation. The framework supports brand owners and designers in designing their products. To prevent misuse, including brand impersonation, its use should be limited to authorized brands, with final decisions made by human designers.

\bibliography{aaai2026}

\onecolumn

\setcounter{section}{0}
\setcounter{figure}{0}
\setcounter{table}{0}
\setcounter{equation}{0}

\renewcommand{\thesection}{S\arabic{section}}
\renewcommand{\thefigure}{S\arabic{figure}}
\renewcommand{\thetable}{S\arabic{table}}
\renewcommand{\theequation}{S\arabic{equation}}

\section*{Supplemental Material}
\addcontentsline{toc}{section}{Supplemental Material}


\section{Analysis of VLM-as-Judge Evaluation} \label{app:vlm_as_judge}

In this section, we analyze the reliability of our VLM-as-Judge evaluation protocol, with a particular focus on whether the scalar Creativity Scores it produces are consistent with alternative evaluation paradigms.

\begin{figure}[htbp]
    \centering
    \includegraphics[width=0.45\textwidth]{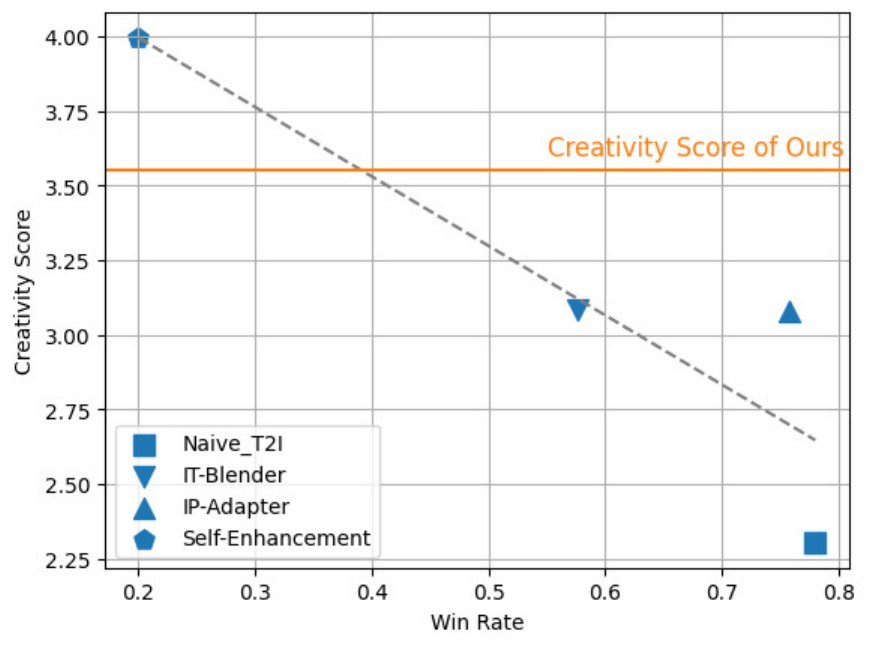}
    \caption{Creativity Score and win rate of the proposed framework against the baselines.}
    \label{fig:win_rate_against_creativity}
\end{figure}


\begin{table}[htbp]
\centering
\begin{tabular}{@{}llll@{}}
\toprule
                 & Gemini  & Prometheus & ArtMuse \\ \midrule
Self-Enhancement & 1 (4.0) & 1 (3.8)    & 1 (63)  \\
Ours             & 2 (3.5) & 3 (3.7)    & 2 (62)  \\
IT-Blender       & 3 (3.1) & 2 (3.7)    & 4 (60)  \\
IP-Adapter       & 4 (3.1) & 4 (3.6)    & 5 (60)  \\
Naive T2I        & 5 (2.3) & 5 (3.6)    & 3 (62)  \\ \bottomrule
\end{tabular}
\caption{Rank of generation approaches by the Creativity Scores of our evaluation protocol and previous VLM-based evaluators, Prometheus-Vision and ArtiMuse. The number in parentheses is the Creativity Score of each approach.}
\label{table:creativity_rank_comparison}
\end{table}


\textbf{Consistency with pairwise comparison: }
We first examine whether the Creativity Scores align with preferences obtained from direct pairwise comparison, which is a commonly used evaluation method for subjective qualities such as creativity.
To this end, we construct a direct-comparison setting in which the VLM is given a pair of images, one generated by the proposed framework and one by a baseline, and asked to select the more creative image. 
From these comparisons, we compute the win rate of the proposed framework against each baseline.

Figure~\ref{fig:win_rate_against_creativity} plots Creativity Score as a function of the win rate. 
We observe a strong monotonic relationship: the proposed framework achieves higher win rates against baselines with lower Creativity Scores, and lower win rates against those with higher Creativity Scores. 
This result indicates that the scalar Creativity Scores produced by our evaluation protocol are well aligned with pairwise preferences. 
Therefore, the proposed scoring scheme provides a reliable proxy for direct comparison while being more scalable.

\textbf{Robustness across different VLM-based evaluators: }
We next evaluate whether our conclusions depend on the specific choice of VLM or prompting strategy.
To this end, we compare our evaluation results with those obtained using two prior VLM-based evaluators, Prometheus-Vision~\cite{lee-etal-2024-prometheus} and ArtiMuse~\cite{cao2025artimusefinegrainedimageaesthetics}. 
Table~\ref{table:creativity_rank_comparison} reports the resulting Creativity Scores and rankings.

Although the exact rankings differ slightly across evaluators, all methods consistently identify Self-Enhancement as the most creative approach, followed by the proposed framework. 
This agreement is consistent with our main result, where Self-Enhancement achieves the highest Creativity Score at the cost of a lower Identity Score, while our framework attains the second-highest Creativity Score with substantially better brand identifiability.
These results support our main conclusions, which are robust to the choice of evaluator and are not an artifact of a specific VLM or prompt design.

\section{Prompt Templates Used in the Implementation} \label{app:survey_prompt}

This appendix lists the main prompt templates used in the implementation. We report representative prompt templates and preserve the original placeholders used in the code.

The brand image survey report is constructed with four category-specific prompts (research details, market analysis, target customer, and customer value), followed by a category-independent summarization prompt. The category-specific templates are listed below.

\begin{table}[h]
\small
\centering
\begin{tabular}{|p{0.18\textwidth}|p{0.76\textwidth}|}
\hline
Prompt Name & Template \\
\hline
Research details & You are an automotive market analyst.\newline\newline Research and summarize a comprehensive product overview of \{item\_name\}.\newline\newline Your analysis should include:\newline - Product positioning and purpose\newline - Key specifications and technologies\newline - Unique features and differentiators\newline - Typical use cases and ownership context\newline - Historical background and evolution (if relevant)\newline\newline Write objectively and factually, using clear business-oriented language. \\
\hline
Market analysis & You are conducting a professional market analysis for an automotive product.\newline\newline Based on the following product overview:\newline \{item\_inf\}\newline\newline Analyze the market for \{item\_name\}, covering:\newline - Market size and growth trends\newline - Competitive landscape (key competitors and alternatives)\newline - Market segmentation (e.g., price, technology, usage)\newline - Regulatory, technological, and societal trends affecting demand\newline - Strengths and weaknesses of \{item\_name\} relative to competitors\newline\newline Focus on insights that are relevant for strategic decision-making. \\
\hline
Target customer & You are analyzing target customers for an automotive product.\newline\newline Using the following information:\newline - Product overview: \{item\_inf\}\newline - Market analysis: \{market\_inf\}\newline\newline Identify and describe the primary and secondary target customer segments for \{item\_name\}.\newline\newline For each segment, include:\newline - Demographic characteristics\newline - Psychographic traits (values, attitudes, lifestyle)\newline - Mobility needs and driving behavior\newline - Purchase motivations and constraints\newline - Why this product is relevant to them\newline\newline Present the segments clearly and logically. \\
\hline
Customer value & You are tasked with identifying the key customer value delivered by an automotive product.\newline\newline Based on the following inputs:\newline - Product overview: \{item\_inf\}\newline - Market analysis: \{market\_inf\}\newline - Target customers: \{target\_inf\}\newline\newline Identify and articulate the core customer values that \{item\_name\} provides.\newline\newline Your analysis should:\newline - Link product features to customer needs and pain points\newline - Distinguish functional, emotional, and social value\newline - Explain why these values matter more than competing alternatives\newline - Highlight the primary value proposition for each key customer segment\newline\newline Use clear, structured, and value-oriented language. \\
\hline
\end{tabular}
\caption{Survey report construction prompt templates for the \texttt{automotive} category}
\end{table}

\begin{table}[h]
\small
\centering
\begin{tabular}{|p{0.18\textwidth}|p{0.76\textwidth}|}
\hline
Prompt Name & Template \\
\hline
Research details & You are a two-wheeler and lifestyle mobility market analyst.\newline\newline Research and summarize a comprehensive product overview of \{item\_name\}.\newline\newline Your analysis should include:\newline - Brand positioning and cultural meaning\newline - Engine type, performance characteristics, and core technologies\newline - Design philosophy and signature styling elements\newline - Riding experience and typical usage scenarios (e.g., leisure, touring, commuting)\newline - Historical background, heritage, and model lineage (if relevant)\newline\newline Write objectively and factually, using clear business-oriented language. \\
\hline
Market analysis & You are conducting a professional market analysis for a motorcycle product.\newline\newline Based on the following product overview:\newline \{item\_inf\}\newline\newline Analyze the market for \{item\_name\}, covering:\newline - Market size and growth trends in the motorcycle segment\newline - Competitive landscape (direct competitors and substitute mobility options)\newline - Market segmentation (e.g., displacement, riding style, lifestyle orientation)\newline - Regulatory, demographic, and cultural trends affecting demand\newline - Strengths and weaknesses of \{item\_name\} relative to competing models\newline\newline Focus on insights relevant to strategic and brand decision-making. \\
\hline
Target customer & You are analyzing target customers for a motorcycle product.\newline\newline Using the following information:\newline - Product overview: \{item\_inf\}\newline - Market analysis: \{market\_inf\}\newline\newline Identify and describe the primary and secondary target customer segments for \{item\_name\}.\newline\newline For each segment, include:\newline - Demographics (age, income, riding experience)\newline - Psychographics (identity, lifestyle, values)\newline - Riding needs and usage patterns\newline - Emotional and symbolic purchase motivations\newline - Reasons this motorcycle resonates with them\newline\newline Present the segments clearly and logically. \\
\hline
Customer value & You are tasked with identifying the key customer value delivered by a motorcycle product.\newline\newline Based on the following inputs:\newline - Product overview: \{item\_inf\}\newline - Market analysis: \{market\_inf\}\newline - Target customers: \{target\_inf\}\newline\newline Identify and articulate the core customer values that \{item\_name\} provides.\newline\newline Your analysis should:\newline - Link riding experience and design to customer desires\newline - Distinguish functional, emotional, and identity-related value\newline - Explain how the product delivers lifestyle or self-expression benefits\newline - Highlight the dominant value proposition for each key customer segment\newline\newline Use clear, structured, and value-oriented language. \\
\hline
\end{tabular}
\caption{Survey report construction prompt templates for the \texttt{motorcycle} category}
\end{table}

\begin{table}[h]
\small
\centering
\begin{tabular}{|p{0.18\textwidth}|p{0.76\textwidth}|}
\hline
Prompt Name & Template \\
\hline
Research details & You are a consumer packaged goods (CPG) and beverage market analyst.\newline\newline Research and summarize a comprehensive product overview of \{item\_name\}.\newline\newline Your analysis should include:\newline - Brand positioning and core promise\newline - Product formulation and key attributes\newline - Packaging, variants, and distribution formats\newline - Consumption occasions and usage contexts\newline - Brand history and evolution (if relevant)\newline\newline Write objectively and factually, using clear business-oriented language. \\
\hline
Market analysis & You are conducting a professional market analysis for a beverage product.\newline\newline Based on the following product overview:\newline \{item\_inf\}\newline\newline Analyze the market for \{item\_name\}, covering:\newline - Market size and category growth trends\newline - Competitive landscape and substitute products\newline - Market segmentation (e.g., age, lifestyle, consumption occasion)\newline - Health, regulatory, and cultural trends affecting demand\newline - Strengths and weaknesses of \{item\_name\} relative to competitors\newline\newline Focus on insights relevant to brand and portfolio strategy. \\
\hline
Target customer & You are analyzing target customers for a beverage product.\newline\newline Using the following information:\newline - Product overview: \{item\_inf\}\newline - Market analysis: \{market\_inf\}\newline\newline Identify and describe the primary and secondary target customer segments for \{item\_name\}.\newline\newline For each segment, include:\newline - Demographic profile\newline - Lifestyle and consumption behavior\newline - Emotional and situational consumption drivers\newline - Purchase motivations and barriers\newline - Why the brand is meaningful or relevant to them\newline\newline Present the segments clearly and logically. \\
\hline
Customer value & You are tasked with identifying the key customer value delivered by a beverage product.\newline\newline Based on the following inputs:\newline - Product overview: \{item\_inf\}\newline - Market analysis: \{market\_inf\}\newline - Target customers: \{target\_inf\}\newline\newline Identify and articulate the core customer values that \{item\_name\} provides.\newline\newline Your analysis should:\newline - Link product attributes to consumption needs and emotions\newline - Distinguish functional (taste, refreshment), emotional, and social value\newline - Explain how the brand creates habitual or symbolic value\newline - Highlight the primary value proposition for each customer segment\newline\newline Use clear, structured, and value-oriented language. \\
\hline
\end{tabular}
\caption{Survey report construction prompt templates for the \texttt{beverage} category}
\end{table}

\begin{table}[h]
\small
\centering
\begin{tabular}{|p{0.18\textwidth}|p{0.76\textwidth}|}
\hline
Prompt Name & Template \\
\hline
Research details & You are an entertainment and IP (intellectual property) market analyst.\newline\newline Research and summarize a comprehensive product overview of the character series \{item\_name\}.\newline\newline Your analysis should include:\newline - Core concept and narrative universe\newline - Main characters and thematic elements\newline - Media formats and platforms (e.g., film, games, merchandise)\newline - Audience engagement and usage contexts\newline - Brand history and expansion strategy (if relevant)\newline\newline Write objectively and factually, using clear business-oriented language. \\
\hline
Market analysis & You are conducting a professional market analysis for an entertainment IP.\newline\newline Based on the following product overview:\newline \{item\_inf\}\newline\newline Analyze the market for \{item\_name\}, covering:\newline - Market size and growth of the relevant entertainment segments\newline - Competitive landscape (other franchises and IPs)\newline - Audience and platform segmentation\newline - Technological and cultural trends affecting engagement\newline - Strengths and weaknesses of \{item\_name\} relative to competing IPs\newline\newline Focus on insights relevant to IP strategy and long-term value creation. \\
\hline
Target customer & You are analyzing target audiences for an entertainment character series.\newline\newline Using the following information:\newline - Product overview: \{item\_inf\}\newline - Market analysis: \{market\_inf\}\newline\newline Identify and describe the primary and secondary target audience segments for \{item\_name\}.\newline\newline For each segment, include:\newline - Age group and life stage\newline - Emotional and psychological appeal\newline - Engagement behavior across media and products\newline - Motivations for attachment and loyalty\newline - Why the characters or world resonate with them\newline\newline Present the segments clearly and logically. \\
\hline
Customer value & You are tasked with identifying the key audience value delivered by an entertainment character series.\newline\newline Based on the following inputs:\newline - Product overview: \{item\_inf\}\newline - Market analysis: \{market\_inf\}\newline - Target audiences: \{target\_inf\}\newline\newline Identify and articulate the core values that \{item\_name\} provides to its audience.\newline\newline Your analysis should:\newline - Link storytelling, characters, and world-building to emotional needs\newline - Distinguish entertainment, emotional, social, and nostalgic value\newline - Explain why the series sustains long-term engagement\newline - Highlight the primary value proposition for each audience segment\newline\newline Use clear, structured, and value-oriented language. \\
\hline
\end{tabular}
\caption{Survey report construction prompt templates for the \texttt{character\_series} category}
\end{table}

\begin{table}[h]
\small
\centering
\begin{tabular}{|p{0.18\textwidth}|p{0.76\textwidth}|}
\hline
Prompt Name & Template \\
\hline
Research details & You are an automotive industry and corporate strategy analyst.\newline\newline Research and summarize a comprehensive overview of the automobile manufacturer \{item\_name\}.\newline\newline Your analysis should include:\newline - Corporate positioning and brand philosophy\newline - Product portfolio and technology strategy\newline - Manufacturing, quality, and supply chain characteristics\newline - Global presence and key markets\newline - Corporate history and strategic evolution\newline\newline Write objectively and factually, using clear business-oriented language. \\
\hline
Market analysis & You are conducting a professional market analysis for an automobile manufacturer.\newline\newline Based on the following overview:\newline \{item\_inf\}\newline\newline Analyze the market position of \{item\_name\}, covering:\newline - Global and regional market share trends\newline - Competitive landscape across vehicle segments\newline - Portfolio coverage and customer segmentation\newline - Regulatory, technological, and mobility trends affecting the company\newline - Strengths and weaknesses of \{item\_name\} relative to major competitors\newline\newline Focus on insights relevant to corporate and long-term strategy. \\
\hline
Target customer & You are analyzing target customers for an automobile manufacturer.\newline\newline Using the following information:\newline - Company overview: \{item\_inf\}\newline - Market analysis: \{market\_inf\}\newline\newline Identify and describe the major customer segments served by \{item\_name\}.\newline\newline For each segment, include:\newline - Demographic and geographic characteristics\newline - Mobility needs and usage patterns\newline - Brand perception and expectations\newline - Purchase motivations and decision criteria\newline - Why this manufacturer appeals to them\newline\newline Present the segments clearly and logically. \\
\hline
Customer value & You are tasked with identifying the key customer value delivered by an automobile manufacturer.\newline\newline Based on the following inputs:\newline - Company overview: \{item\_inf\}\newline - Market analysis: \{market\_inf\}\newline - Target customers: \{target\_inf\}\newline\newline Identify and articulate the core customer values that \{item\_name\} delivers across its portfolio.\newline\newline Your analysis should:\newline - Link corporate capabilities and product strengths to customer needs\newline - Distinguish functional, emotional, and trust-based value\newline - Explain how the brand sustains long-term customer relationships\newline - Highlight the primary value propositions for key customer segments\newline\newline Use clear, structured, and value-oriented language. \\
\hline
\end{tabular}
\caption{Survey report construction prompt templates for the \texttt{automobile\_manufacturer} category}
\end{table}

\begin{table}[h]
\small
\centering
\begin{tabular}{|p{0.18\textwidth}|p{0.76\textwidth}|}
\hline
Prompt Name & Template \\
\hline
Summary & Summarize the following product report in 500 words:\newline\newline \{full\_report\} \\
\hline
\end{tabular}
\caption{Category-independent summarization prompt used to compress the survey report}
\end{table}

\begin{table}[h]
\small
\centering
\begin{tabular}{|p{0.18\textwidth}|p{0.76\textwidth}|}
\hline
Prompt Name & Template \\
\hline
Image concept & Find the concept/message/tone of the image. \\
\hline
Direction integration & This is a market report/target customer/customer value of \{model\_name\}.\newline\newline Market report:\newline \{market\_report\}\newline\newline Target customer:\newline \{target\_customer\}\newline\newline Customer values:\newline \{customer\_values\}\newline\newline Now you integrate the concept/message/tone of the image into the \{model\_name\} design. Explain how the concept/message/tone of the image can be integrated into the visual context of new \{model\_name\} without losing its essence. \\
\hline
Direction guideline & Write a guideline to evaluate how much the exterior of the new \{model\_name\} aligns with the concept/message/tone by 1-10 in one paragraph. \\
\hline
Direction selection & This is a market report/target customer/customer value of \{model\_name\}.\newline\newline Market report:\newline \{market\_report\}\newline\newline Target customer:\newline \{target\_customer\}\newline\newline Customer values:\newline \{customer\_values\}\newline\newline Here are candidates of design directions for the new \{model\_name\}:\newline\newline \{design\_directions\}\newline\newline Select the best design direction that effectively appeals to the target customers based on the market report, target customer, and customer values. Decide by ORIGINALITY, NOVELTY, SUPRISENESS and CLEARNESS of the design direction. Return just one alphabet corresponding to the best design direction. \\
\hline
\end{tabular}
\caption{Creative direction prompt templates}
\end{table}

\begin{table}[h]
\small
\centering
\begin{tabular}{|p{0.18\textwidth}|p{0.76\textwidth}|}
\hline
Prompt Name & Template \\
\hline
Initial prompt & You are designing the new \{model\_name\}:\newline \{short\_summary\}\newline\newline Write a prompt for a stable diffusion model for the new \{model\_name\} respecting the important characteristics in the above description. No background is allowed in the image. \\
\hline
Prompt refinement & Read the description about \{model\_name\}:\newline\newline \{short\_summary\}\newline\newline Impvoe the following prompt to generate the image of the new \{model\_name\} by reflecting the feedback with keeping the important characteristics in the above description.\newline\newline Current Prompt:\newline \{current\_prompt\}\newline\newline Feedback:\newline \{engineer\_feedback\}\newline\newline You propose a strategy to plan how you will apply the feedback to enhance the prompt. Then, you rewrite the prompt accordingly. Prompts must be in 50 words.\newline Provide your response in the specified JSON format. \{"strategy": "...", "prompt": "..."\} \\
\hline
Design review & You are acting as an expert design reviewer evaluating an image of the new \{model\_name\} generated by a diffusion model.\newline\newline You are provided with the following contextual information to guide your evaluation:\newline\newline Market Report:\newline \{market\_report\}\newline\newline Target Customer:\newline \{target\_customer\}\newline\newline Customer Values:\newline \{customer\_values\}\newline\newline Using this context, evaluate the image of new \{model\_name\}.\newline\newline Score the image according to the criteria listed below:\newline\newline \{score\_metrics\}\newline\newline You make a comment on the work by each criteria and give its score from 1 to 10. Provide your response in the specified JSON format. \{json\_format\} \\
\hline
\end{tabular}
\caption{Self-enhancement prompt templates for the Designer (initial prompt and prompt refinement) and the Reviewer (design review)}
\end{table}

\begin{table}[h]
\small
\centering
\begin{tabular}{|p{0.18\textwidth}|p{0.76\textwidth}|}
\hline
Criterion & Definition \\
\hline
Originality & Originality in creativity centers on producing novel and unique ideas. Margaret Boden defines creativity as the ability to generate outcomes that are ``new, surprising, and valuable,'' emphasizing novelty as a core criterion. Her theory differentiates combinational, exploratory, and transformational creativity -- all mechanisms for breaking beyond existing concepts to achieve originality. J. P. Guilford's divergent thinking framework likewise highlights originality: creative thinkers produce many varied solutions, with originality measured by the rarity and uniqueness of ideas. Together, these theories establish that genuine creativity requires ideas that depart from the ordinary and expand conceptual boundaries. \\
\hline
Unexpected\-Associations & This principle involves making surprising connections between disparate ideas -- a key source of creative ingenuity. Finke, Ward, and Smith's Geneplore model of creative cognition describes a two-stage process: a generation phase where preliminary ideas or mental structures are produced, and an exploration phase where those structures are elaborated and interpreted in novel ways. Through this process, creators often discover unforeseen linkages between concepts, leading to inventive combinations. Margaret Boden's concept of combinational creativity similarly refers to creativity that arises by synthesizing familiar ideas into new, unexpected configurations. Both frameworks stress that many creative breakthroughs happen by recombining knowledge in unconventional ways, yielding associations that surprise because they deviate from usual patterns while still making intuitive sense in hindsight. \\
\hline
CustomerValues & The new design must effectively reflect and embody the customer values that are important to the target customers described above. Each design element should be thoughtfully chosen to resonate with these values, ensuring that the overall aesthetic and functionality align with what the customers prioritize and appreciate. \\
\hline
\end{tabular}
\caption{Criterion definitions inserted as \texttt{\{score\_metrics\}} into the design review prompt of the Reviewer}
\end{table}

\begin{table}[h]
\small
\centering
\begin{tabular}{|p{0.18\textwidth}|p{0.76\textwidth}|}
\hline
Category & Format criterion definition \\
\hline
Automotive & The image must NOT have any other object except the car. The image must shows the car from a front-left viewpoint. The entire car must be fully visible in the image. No text or watermark should be present in the image. \\
\hline
Motorcycle & The image must NOT have any other object except the motorcycle. The image must shows the motorcycle from a front-left viewpoint. The entire motorcycle must be fully visible in the image. No text or watermark should be present in the image. \\
\hline
Beverage & The image must NOT have any other object except the labeled beverage can or bottle. No text or watermark should be present in the image. \\
\hline
Character series & The image must NOT have any other object except the character. No text or watermark should be present in the image. \\
\hline
Automobile manufacturer & The image must NOT have any other object except the logo. No text or watermark should be present in the image. \\
\hline
\end{tabular}
\caption{Category-specific \texttt{Format} criterion definitions inserted as part of \texttt{\{score\_metrics\}} into the design review prompt of the Reviewer}
\end{table}

\begin{table}[h]
\small
\centering
\begin{tabular}{|p{0.18\textwidth}|p{0.76\textwidth}|}
\hline
Category & Template \\
\hline
Automotive & You are given an image that may depict an automobile or something resembling one. Based only on visual features such as body shape, headlights, grille, proportions, styling era, and brand design language, predict the most likely automobile product name.\newline\newline Rules:\newline - Output exactly one specific car model name (make + model if applicable).\newline - You must give a best guess even if the image is unclear.\newline - Do not answer with generic terms  (e.g., "sedan", "SUV").\newline - Do not add explanations or qualifiers.\newline\newline Output format:\newline Automotive: <best guess automobile model> \\
\hline
Motorcycle & You are given an image that may depict a motorcycle or motorcycle-related object. Using visual cues such as frame geometry, engine layout, exhaust placement, fairing style, and brand-specific design language, predict the most likely motorcycle model name.\newline\newline Rules:\newline - Output exactly one specific motorcycle model name.\newline - You must provide a best guess even if uncertain.\newline - Avoid generic classes (e.g., "sport bike", "cruiser").\newline - No explanations or disclaimers.\newline\newline Output format:\newline Motorcycle: <best guess motorcycle model> \\
\hline
Beverage & You are given an image that may depict a beverage or beverage container. Based on packaging shape, label design, color scheme, typography, and common market products, predict the most likely beverage product name.\newline\newline Rules:\newline - Output exactly one concrete beverage product name.\newline - Always give a best guess, even if the image is ambiguous.\newline - Do not output beverage categories (e.g., "soda", "energy drink").\newline - Do not include explanations.\newline\newline Output format:\newline Beverage: <best guess beverage product> \\
\hline
\end{tabular}
\caption{VLM-as-Judge prompts for identity evaluation I}
\end{table}

\begin{table}[h]
\small
\centering
\begin{tabular}{|p{0.18\textwidth}|p{0.76\textwidth}|}
\hline
Category & Template \\
\hline
Character series & You are given an image that may depict a fictional character. Using visual traits such as clothing, hairstyle, facial features, accessories, color palette, and art style, predict the most likely character series or franchise the character belongs to.\newline\newline Rules:\newline - Output exactly one character series or franchise name.\newline - Do NOT output the individual character name.\newline - You must choose the single most likely series, even if uncertain.\newline - Do not say ``original character'' or ``unknown''.\newline - No explanations or alternatives.\newline\newline Output format:\newline Character series: <series name> \\
\hline
Automobile manufacturer & You are given an image that may depict a logo or emblem. Based on shape, symbols, typography, color usage, and known automotive branding patterns, predict the most likely automobile manufacturer.\newline\newline Rules:\newline - Output exactly one automobile manufacturer name.\newline - Always provide a best guess, even if the logo is unclear or stylized.\newline - Do not list multiple candidates.\newline - Do not include explanations.\newline\newline Output format:\newline Automobile manufacturer logo: <manufacturer name> \\
\hline
\end{tabular}
\caption{VLM-as-Judge prompts for identity evaluation II}
\end{table}

\begin{table}[h]
\small
\centering
\begin{tabular}{|p{0.18\textwidth}|p{0.76\textwidth}|}
\hline
Category & Template \\
\hline
Automotive & You are given an image of a newly designed \{item\_name\}.\newline\newline Evaluate the creativity of the vehicle's exterior visual design as depicted in the image.\newline\newline Evaluation guidelines (must follow strictly):\newline - Assess features such as overall silhouette, body proportions, surface treatment, lighting signatures, grille design, and visual coherence.\newline - Focus on how distinctive and imaginative the exterior design appears compared to contemporary automobiles.\newline - Do not evaluate performance, specifications, interior features, or brand reputation.\newline\newline Evaluate the creativity of the visual design based on the following criteria:\newline\newline \{score\_metrics\}\newline\newline You make a comment on the design by each criteria and give its score from 1 to 5. Provide your response in the specified JSON format. \{json\_format\} \\
\hline
Motorcycle & You are given an image of a newly designed \{item\_name\}.\newline\newline Evaluate the creativity of the motorcycle's visual design as depicted in the image.\newline\newline Evaluation guidelines (must follow strictly):\newline - Assess features such as frame layout, bodywork, tank shape, seat profile, lighting, exhaust placement, and overall visual balance.\newline - Focus on how distinctive and imaginative the design appears within the motorcycle category.\newline - Do not evaluate engine performance, riding dynamics, or brand reputation.\newline\newline Evaluate the creativity of the visual design based on the following criteria:\newline\newline \{score\_metrics\}\newline\newline You make a comment on the design by each criteria and give its score from 1 to 5. Provide your response in the specified JSON format.\{json\_format\} \\
\hline
Beverage & You are given an image of a newly designed \{item\_name\}.\newline\newline Evaluate the creativity of the beverage's packaging and visual identity as depicted in the image.\newline\newline Evaluation guidelines (must follow strictly):\newline - Assess features such as container shape, label graphics, typography, color palette, material impression, and shelf presence.\newline - Focus on how distinctive and imaginative the packaging appears compared to existing beverage products.\newline - Do not evaluate taste, ingredients, nutritional value, or brand popularity.\newline\newline Evaluate the creativity of the visual design based on the following criteria:\newline\newline \{score\_metrics\}\newline\newline You make a comment on the design by each criteria and give its score from 1 to 5. Provide your response in the specified JSON format. \{json\_format\} \\
\hline
\end{tabular}
\caption{VLM-as-Judge prompts for creativity evaluation I}
\end{table}

\begin{table}[h]
\small
\centering
\begin{tabular}{|p{0.18\textwidth}|p{0.76\textwidth}|}
\hline
Category & Template \\
\hline
Character series & You are given an image of a newly designed character belonging to a fictional character series: \{item\_name\}.\newline\newline Evaluate the creativity of the character's visual design as depicted in the image.\newline\newline Evaluation guidelines (must follow strictly):\newline - Assess features such as silhouette, costume, color palette, facial design, accessories, and iconic visual traits.\newline - Focus on how distinctive and imaginative the character design appears within animated or fictional media.\newline - Do not evaluate story quality, personality, or popularity of the series.\newline\newline Evaluate the creativity of the visual design based on the following criteria:\newline\newline \{score\_metrics\}\newline\newline You make a comment on the design by each criteria and give its score from 1 to 5. Provide your response in the specified JSON format. \{json\_format\} \\
\hline
Automobile manufacturer & You are given an image of a newly designed automobile manufacturer logo: \{item\_name\}.\newline\newline Evaluate the creativity of the logo's visual design as depicted in the image.\newline\newline Evaluation guidelines (must follow strictly):\newline - Assess features such as shape, symbolism, typography, color usage, simplicity, and visual balance.\newline - Focus on how distinctive and imaginative the logo appears compared to existing automobile logos.\newline - Do not evaluate brand history, reputation, or market success.\newline\newline Evaluate the creativity of the visual design based on the following criteria:\newline\newline \{score\_metrics\}\newline\newline You make a comment on the design by each criteria and give its score from 1 to 5. Provide your response in the specified JSON format. \{json\_format\} \\
\hline
\end{tabular}
\caption{VLM-as-Judge prompts for creativity evaluation II}
\end{table}

\begin{table}[h]
\small
\centering
\begin{tabular}{|p{0.18\textwidth}|p{0.76\textwidth}|}
\hline
Metric & Definition \\
\hline
Novelty & Evaluate the degree to which the product, character, label, or logo depicted in the image appears visually new, uncommon, and differentiated relative to the current (baseline) design, based solely on observable features. The assessment should focus on form, silhouette, composition, proportions, visual language, and stylistic choices, not on technical performance or functionality. Low-fidelity artifacts and stylistic abstraction are acceptable. Novelty must be supported by clear visual divergence, not subtle tweaks or conceptual claims.\newline\newline
1 --- No Novelty (Current Design / Less Novel)\newline
Visually equivalent to the current design or more conservative\newline
Differences, if any, are negligible or purely cosmetic\newline
Overall impression is familiar, expected, or generic\newline
Could be mistaken for the existing design at a glance\newline\newline
2 --- Minimal Novelty\newline
Small, incremental visual changes are present\newline
Variations do not meaningfully alter the overall impression\newline
Design remains largely conventional and unsurprising\newline
Reads as a minor update or derivative variation\newline\newline
3 --- Moderate Novelty\newline
Clear visual differences from the current design\newline
Changes affect some visible aspects (shape, layout, style)\newline
Overall concept remains familiar or within the same design family\newline
Feels like a standard evolution rather than a rethink\newline\newline
4 --- High Novelty\newline
Design is clearly distinct and visually unexpected\newline
Noticeable departure in form, styling, or visual language\newline
Maintains recognizability of purpose while reimagining appearance\newline
Would stand out among existing designs in the same category\newline\newline
5 --- Extreme Novelty\newline
Design is highly unconventional and visually striking\newline
Strong departure from existing norms, patterns, or archetypes\newline
Immediate impression of originality and surprise\newline
Unlikely to be confused with existing designs or trends \\
\hline
\end{tabular}
\caption{Score metric definition inserted as \texttt{\{score\_metrics\}} into the creativity evaluation prompts I: Novelty Score}
\end{table}

\begin{table}[h]
\small
\centering
\begin{tabular}{|p{0.18\textwidth}|p{0.76\textwidth}|}
\hline
Metric & Definition \\
\hline
Usefulness & Evaluate the degree to which the object, product, logo/label, or character depicted in the image appears functionally plausible, logically constructed, and realistically usable based only on visible elements. The assessment should judge whether the design could reasonably be produced, deployed, or appear on-screen as-is, without relying on hidden mechanisms, off-screen explanations, or idealized execution not supported by the image. Low-fidelity rendering, stylization, or AI artifacts are acceptable. However, visible functional contradictions, structural impossibilities, or incoherent design decisions should reduce the score. For existing real-world products or established characters, score 5 if they appear intact and recognizable.\newline\newline
1 --- Non-Functional / Implausible\newline
Design does not make practical or logical sense as depicted\newline
Contains clear impossibilities, contradictions, or incoherent elements\newline
Cannot reasonably be produced, used, or appear on-screen in its current form\newline\newline
2 --- Mostly Non-Functional / Conceptually Flawed\newline
Major structural, logical, or usability issues are visible\newline
Core function is compromised or unclear\newline
Design relies on unrealistic assumptions not supported by the image\newline
Would require fundamental redesign to function or appear credibly\newline\newline
3 --- Functionally Plausible but Underspecified\newline
Overall concept makes sense for its intended purpose or role\newline
Key functional details are unclear, stylized, or incomplete\newline
No direct contradictions, but realism or usability is uncertain\newline
Would require significant clarification or development to be production- or screen-ready\newline\newline
4 --- Largely Functional with Minor Uncertainty\newline
Design is logically constructed and clearly intended to function\newline
Minor ambiguities, missing details, or simplifications are present\newline
No obvious fatal flaws; issues could be resolved during refinement\newline
Appears close to production-ready or screen-ready\newline\newline
5 --- Fully Functional / Production- or Screen-Ready\newline
Design is coherent, complete, and internally consistent\newline
All visible elements logically support the intended purpose or role\newline
No apparent structural, ergonomic, or logical contradictions\newline
Could reasonably be manufactured, deployed, or appear in a movie/anime without redesign \\
\hline
\end{tabular}
\caption{Score metric definition inserted as \texttt{\{score\_metrics\}} into the creativity evaluation prompts II: Usefulness Score}
\end{table}

\end{document}